\documentclass[a4paper,fleqn]{cas-dc}

\usepackage[numbers]{natbib}
\usepackage{graphicx}
\usepackage{booktabs}
\usepackage{multirow}
\usepackage{amsmath,amssymb,amsfonts}
\usepackage{algorithm}
\usepackage{algorithmic}
\usepackage{textcomp}
\usepackage{xcolor}
\usepackage{subcaption}

\usepackage{graphicx}
\usepackage{float}
\usepackage{tabularx}
\usepackage{booktabs}
\usepackage{multirow}

\def\tsc#1{\csdef{#1}{\textsc{\lowercase{#1}}\xspace}}
\tsc{WGM}
\tsc{QE}

\begin{document}
\let\WriteBookmarks\relax
\def\floatpagepagefraction{1}
\def\textpagefraction{.001}

\shorttitle{IoT and Deep Learning for Detection of \textit{Postelectrotermes militaris}}

\shortauthors{Senevirathna et al.}

\title[mode=title]{Effectiveness of IoT and Deep Learning for Detection and Severity Assessment of \textit{Postelectrotermes militaris} in Tea Plantations}



\author[1]{D.K.C. Senevirathna}
\orcidauthor{0009-0002-2530-4206}{D.K.C. Senevirathna}
\fnmark[1]

\author[2]{A.A.E. Nanayakkara}
\orcidauthor{0009-0001-5217-2942}{A.A.E. Nanayakkara}
\fnmark[1]

\author[1]{H.M.C.K. Kulathunga}
\orcidauthor{0009-0003-1792-5907}{H.M.C.K. Kulathunga}

\author[1]{J.K.D.P. Nadula}
\orcidauthor{0009-0009-3343-831X}{J.K.D.P. Nadula}

\author[1]{R.M. Mapatuna}
\orcidauthor{0009-0004-2468-2154}{R.M. Mapatuna}

\author[1]{Malithi Nawarathne}
\orcidauthor{0009-0002-2489-2421}{Malithi Nawarathne}

\author[3]{Jaliya L. Wijayaraja}
\orcidauthor{0009-0000-5988-1429}{Jaliya L. Wijayaraja}

\author[4]{P.D. Senanayake}
\orcidauthor{0000-0001-7630-1430}{P.D. Senanayake}

\author[5]{Samitha Vidhanaarachchi}
\orcidauthor{0000-0003-0534-2977}{Samitha Vidhanaarachchi}

\author[1]{Kalpani Manathunga}
\orcidauthor{0000-0003-1936-5053}{Kalpani Manathunga}
\cormark[1]
\ead{kalpani.m@sliit.lk}


\affiliation[1]{
    organization={Sri Lanka Institute of Information Technology},
    addressline={New Kandy Road},
    city={Malabe},
    country={Sri Lanka}
}

\affiliation[2]{
    organization={School of Molecular and Life Sciences, Curtin University},
    addressline={Kent Street},
    city={Bentley},
    state={Western Australia},
    country={Australia}
}

\affiliation[3]{
    organization={University of Kelaniya},
    addressline={Kandy Road, Dalugama},
    city={Kelaniya},
    country={Sri Lanka}
}

\affiliation[4]{
    organization={Tea Research Institute of Sri Lanka},
    city={Talawakelle},
    country={Sri Lanka}
}

\affiliation[5]{
    organization={Murdoch University},
    addressline={90 South St, Murdoch},
    city={Perth},
    state={Western Australia},
    country={Australia}
}

\fntext[1]{These authors contributed equally to this work; sorted by seniority.}

\cortext[1]{Corresponding author}

\begin{abstract}
Tea plantations are vulnerable to \textit{Postelectrotermes militaris}, commonly known as the Upcountry Live Wood Termite (ULWT), which can cause substantial damage when infestations remain undetected. This study proposes an IoT-enabled acoustic monitoring framework integrated with deep learning for early detection and severity assessment of ULWT infestations in tea plantations.

\textit{Research Method:} 
Audio signals were captured non-invasively from tea trunks using a high-sensitivity microphone connected to a Raspberry Pi-based IoT device, with geographic coordinates recorded for spatial tracking. After trimming, resampling, and segmentation, 2,000 ten-second samples were obtained, comprising 1,000 healthy and 1,000 infested samples, and divided into 1,600 training, 200 validation, and 200 test samples. The dataset used in this study is publicly available on Kaggle (Senevirathna et al. 2026). Fourier-derived spectrograms trained a CNN for infestation classification and probability estimation. A weighted severity model combined CNN probability, mean acoustic amplitude, and nearby infested plants within 5 m, with geospatial mapping used to visualize infestation distribution.

\textit{Findings and Values:} 
Field trials in a ULWT-affected tea plantation in Pundaluoya demonstrated feasibility under realistic environmental noise. On the held-out test set, the CNN achieved 81.5\% accuracy, 80.6\% precision, 83.0\% recall, 81.8\% F1-score, and 0.819 ROC-AUC. Beyond binary infestation detection, the framework introduced quantitative severity assessment using infestation probability, acoustic amplitude, and nearby infested plants. The resulting severity and geospatial outputs can support plantation managers in identifying high-risk areas, prioritizing field inspections, and implementing more timely and targeted control measures.

\end{abstract}





\maketitle

\section{Introduction}

Tea (\textit{Camellia sinensis} L.) is a globally important beverage crop and a major contributor to Sri Lanka's economy through export revenue and employment (Deka \textit{et al.} 2021, Weerasingha 2022). Termite infestations threaten sustainable production by damaging tea bushes, reducing yields, and increasing recovery costs when detection is delayed (Kidanu \textit{et al.} 2020, Xia \textit{et al.} 2020).

\begin{figure*}[t]
    \centering

    \begin{subfigure}[b]{0.48\textwidth}
        \centering
        \includegraphics[width=\linewidth]{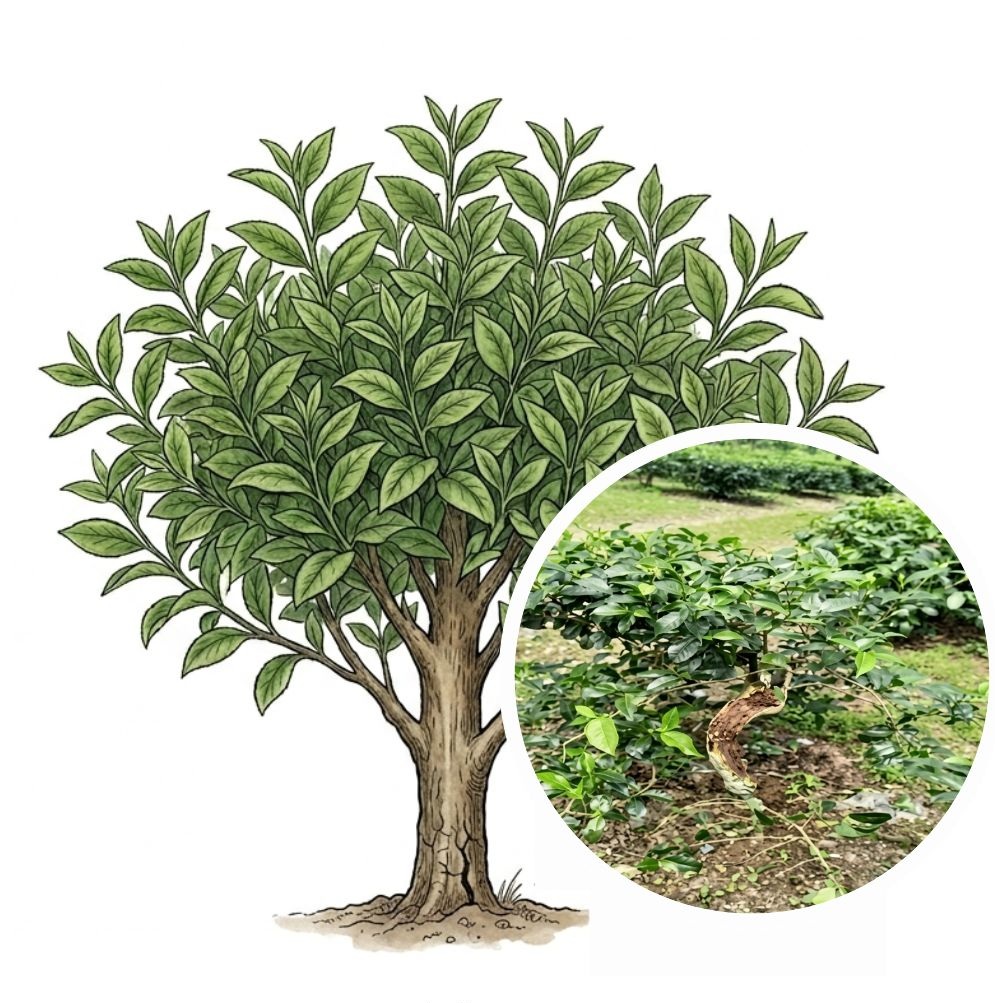}
        \caption{}
        \label{fig:ulwt_a}
    \end{subfigure}
    \hfill
    \begin{subfigure}[b]{0.48\textwidth}
        \centering
        \includegraphics[width=\linewidth]{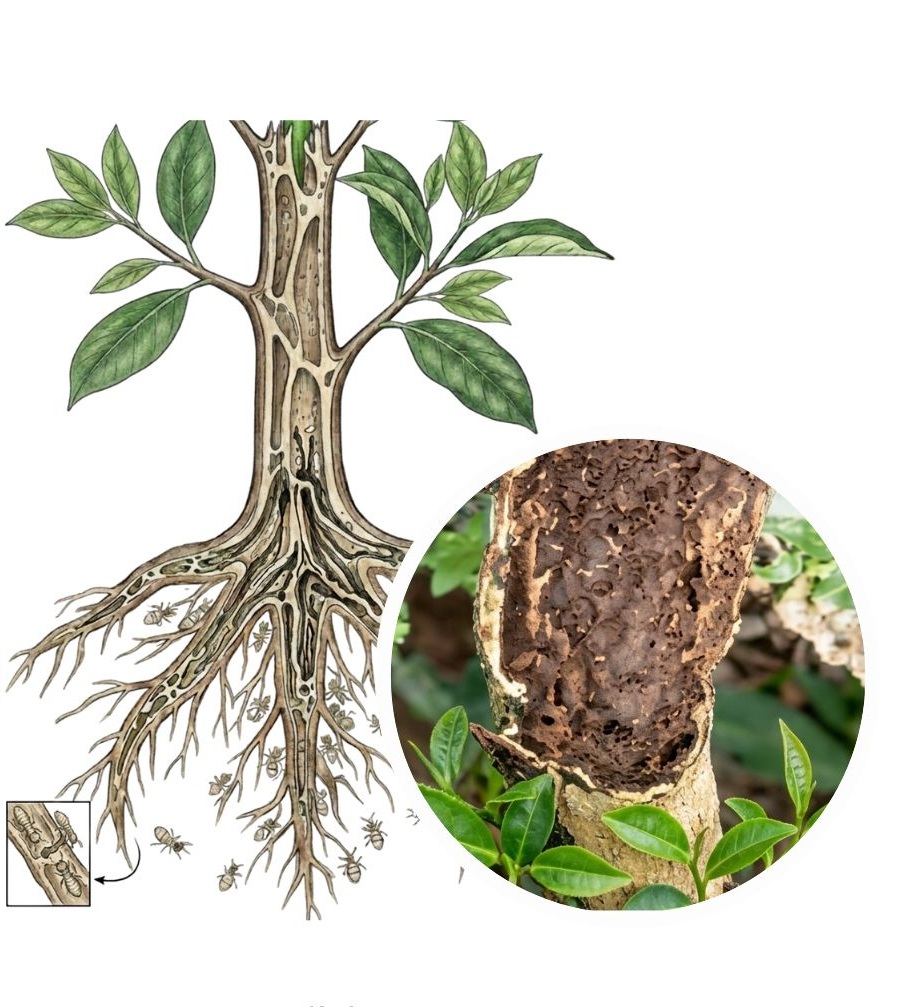}
        \caption{}
        \label{fig:ulwt_b}
    \end{subfigure}

    \caption{(A) Concealed ULWT infestation in a visually healthy tea bush.
    (B) Internal trunk and root damage caused by heartwood galleries.}
    \label{fig:ulwt_damage}
\end{figure*}

The Upcountry Live Wood Termite (ULWT), \textit{Postelectrotermes militaris}, is a major pest of Sri Lankan tea plantations that colonizes living woody tissues. Galleries formed within the heartwood and internal stems progressively weaken plant integrity (Amarasinghe \& Hominick 1993). During early and intermediate infestation, the canopy may remain visually healthy despite substantial internal deterioration (Fig.~\ref{fig:ulwt_damage}), limiting the effectiveness of visual inspection, surface-applied treatments, and natural predation.

\begin{figure}
    \centering
    \includegraphics[width=\linewidth]{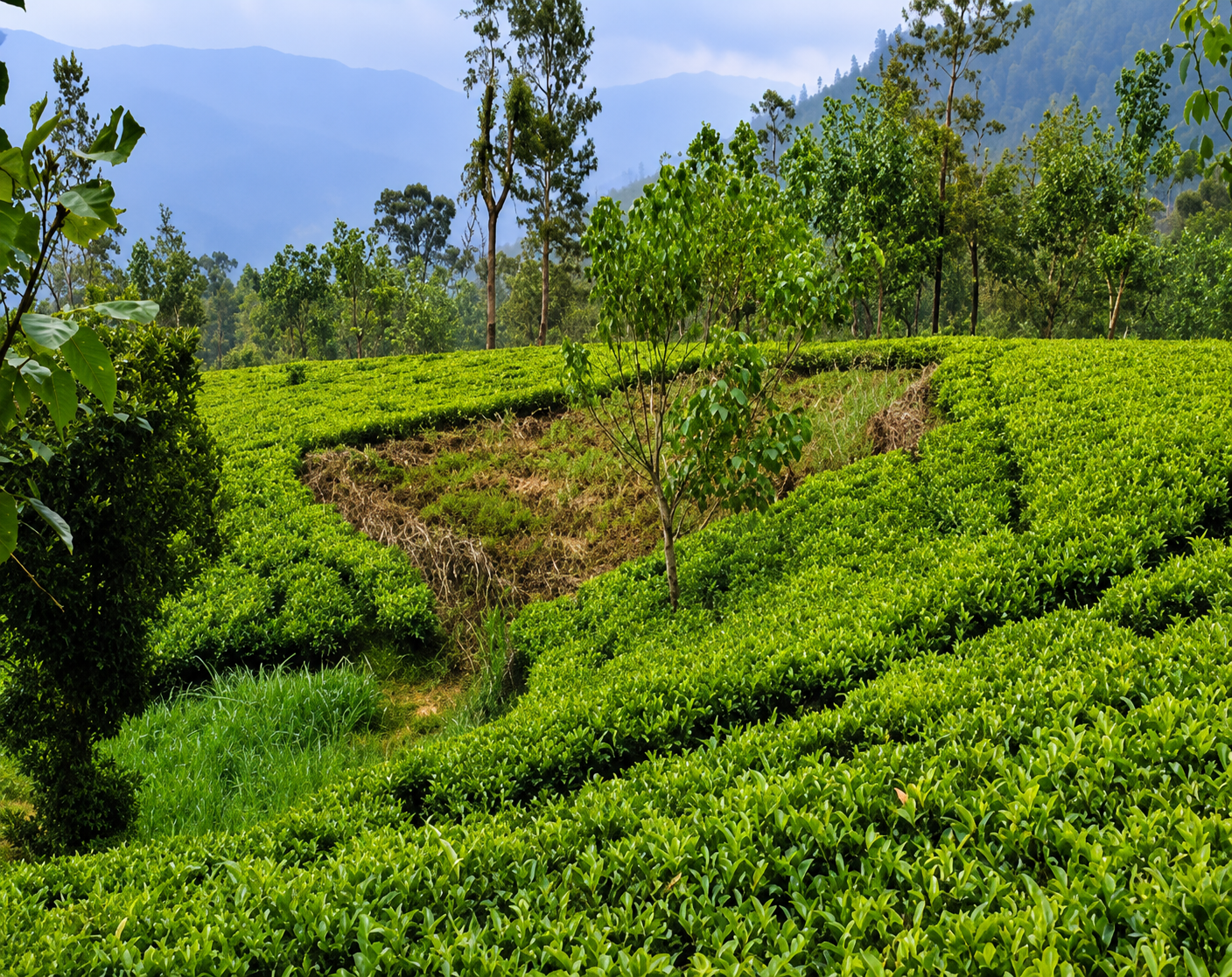}
    \caption{ULWT-affected plantation area following removal of infested tea bushes, showing \textit{Mana} grass (\textit{Glyceria}) cultivation implemented to reduce termite activity before replanting.}
    \label{fig:mana_grass}
\end{figure}

Current ULWT management primarily involves removing and burning infested bushes, although colonies may persist in the surrounding soil and cause reinfestation. Affected areas are therefore commonly replanted with Mana grass (\textit{Glyceria}) before tea cultivation is re-established (Gnanapragasam \textit{et al.} 2018) (Fig.~\ref{fig:mana_grass}). This practice increases labour and maintenance costs and can create a three- to four-year income gap while replanted tea reaches productive maturity (Gnanapragasam \& Sivepalan 2004), limiting its long-term sustainability.

Management is further constrained by labour-intensive inspections that rely heavily on individual expertise. Heavily infested fields can experience substantial stand loss and yield reduction (The Tea Research Institute of Sri Lanka Annual Report 2005), with even moderate yield losses in high-grown plantations causing considerable economic impacts. Effective control therefore requires accurate assessment of infestation severity and spatial distribution, yet current practices lack systematic early-stage monitoring, predictive capability, and spatial tracking (The Tea Research Institute of Sri Lanka Annual Report 2005).

ULWT is particularly prevalent in plantations above 1,000 m elevation (Vitarana \& Mohotti 2008). Termites enter through the root system and progressively colonise internal stems and branches, allowing structural damage and yield decline to develop before visible symptoms appear. These limitations highlight the need for scalable, non-invasive technologies for early detection and localization.

Automated pest monitoring has increasingly used sensor-based systems and machine learning (ML) and deep learning (DL) techniques to reduce reliance on manual inspection (Nanda \textit{et al.} 2019, Escola \textit{et al.} 2020, Ali \textit{et al.} 2023, Dhanaraj \textit{et al.} 2024). Vision-based DL has become a prominent approach for agricultural pest and disease monitoring (Wanninayake \textit{et al.} 2023, Vidhanaarachchi \textit{et al.} 2025).

Convolutional neural networks (CNNs) and object-detection architectures, including YOLO variants, Inception-based models, and MAM-IncNet, have achieved strong performance in visible pest and crop-disease recognition (Chen \textit{et al.} 2024, He \textit{et al.} 2024, Khan \textit{et al.} 2024). Spatial and channel reconstruction mechanisms and attention-based feature refinement have further improved detection robustness (Chithambarathanu \& Jeyakumar 2023, Hewawitharana \textit{et al.} 2023). In tea cultivation, DL has been applied to \textit{Camellia} pest recognition, blister blight detection, and prediction of changing pest and disease risks (Kong \textit{et al.} 2022, Hewawitharana \textit{et al.} 2023, Lee \& Yun 2023, Chen \textit{et al.} 2024).

Representative vision-based studies are summarized in Table~\ref{tab:vision_dl}. Despite their reported performance, these systems depend on externally observable pests or symptoms and are often evaluated under controlled imaging conditions, which may limit generalizability in heterogeneous plantations. Performance may also decline when symptoms are subtle, occluded, or confused with other stresses. More importantly, image-based approaches are unsuitable for concealed ULWT infestation, where internal gallery formation and structural deterioration can occur without visible canopy symptoms.

Acoustic sensing provides a non-destructive approach for detecting concealed insect activity because termites and wood-boring insects generate acoustic emissions during feeding and movement that propagate through woody tissues (Mankin \textit{et al.} 2011, 2020, 2021). Low-cost acoustic systems have demonstrated the feasibility of concealed infestation detection, while subsequent work has advanced digital signal processing and automated classification for pest monitoring (Mankin \textit{et al.} 2020, Bhairavi \textit{et al.} 2020).

\begin{table*}[t]
\centering
\caption{Representative vision-based deep learning systems relevant to pest and disease detection.}
\label{tab:vision_dl}
\renewcommand{\arraystretch}{1.15}

\begin{tabularx}{\textwidth}{
    >{\raggedright\arraybackslash}p{4.5cm}
    >{\raggedright\arraybackslash}p{1.5cm}
    >{\raggedright\arraybackslash}p{2.4cm}
    >{\raggedright\arraybackslash}p{4.8cm}
    >{\raggedright\arraybackslash}X
}
\toprule
\textbf{Ref.} &
\textbf{Crop} &
\textbf{Detection task} &
\textbf{Method} &
\textbf{Performance} \\
\midrule

Kadethankar \textit{et al.} (2021) &
Coconut &
Pest detection &
CNN, Faster R-CNN &
84.64\% \\

Kong \textit{et al.} (2022) &
Multi-crop &
Pest detection &
Deep learning, Fe-Net &
85.29\% \\

Hewawitharana \textit{et al.} (2023) &
Tea &
Disease detection &
CNN, YOLOv8, Mask R-CNN &
98\% recall \\

Wanninayake \textit{et al.} (2023) &
Rice &
Disease \& pest detection &
YOLOv5, CNN, ML &
89--99\% \\

Chen \textit{et al.} (2024) &
Tea &
Pest detection &
MAM-IncNet &
81.44\% recall \\

He \textit{et al.} (2024) &
Tea &
Pest detection &
YOLOv7, SSD, R-CNN &
+2.58\% mAP \\

Khan \textit{et al.} (2024) &
Multi-crop &
Pest detection &
YOLOv5 &
95\% mAP \\

Vidhanaarachchi \textit{et al.} (2025) &
Coconut &
Disease \& pest detection &
CNN, Mask R-CNN, YOLO &
90--97\% \\

\bottomrule
\end{tabularx}
\end{table*}

Recent DL approaches, including CNN--LSTM models, Audio Spectrogram Transformers (ASTs), autoencoders, and semi-supervised learning, have improved discrimination of termite-generated signals from environmental noise, particularly with limited labelled data (Coelho \textit{et al.} 2021, Fagervik \textit{et al.} 2025, Lian \textit{et al.} 2025). Termite feeding, movement, and excavation generate low-amplitude, intermittent vibrations that propagate through woody tissues. These signals vary with colony activity, infestation severity, wood properties, moisture, sensor placement, and environmental conditions, making separation from plantation noise challenging.

Although ASTs capture long-range dependencies, they generally require larger labelled datasets and greater computational resources. CNNs are more efficient for extracting local time-frequency features from spectrograms and are better suited to resource-constrained edge deployment. Given the limited labelled ULWT data and the need for real-time field inference, a CNN-based architecture was selected to balance detection performance, computational efficiency, and deployment feasibility.

Despite promising results, existing acoustic classification methods face several limitations for direct application in tea plantations. Many studies rely on laboratory or semi-controlled recordings with substantially lower background interference than operational fields, where wind, rainfall, machinery, human activity, and non-target organisms may reduce performance. Small or narrowly distributed datasets may also limit generalisability across pest life stages, plant conditions, sensor positions, and environmental settings. Most approaches further focus on binary detection rather than infestation severity or spatial distribution.

Representative acoustic and sensor-based pest-monitoring studies are summarised in Table~\ref{tab:acoustic_sensor}. Although these approaches report promising performance across different pests and agricultural settings, much of the research remains limited to controlled or semi-controlled conditions and does not fully address plantation-scale deployment, environmental interference, or severity assessment.

\begin{table*}[t]
\centering
\caption{Representative acoustic and sensor-based approaches for pest infestation detection.}
\label{tab:acoustic_sensor}
\renewcommand{\arraystretch}{1.15}

\begin{tabularx}{\textwidth}{
    >{\raggedright\arraybackslash}p{4.0cm}
    >{\raggedright\arraybackslash}p{2.3cm}
    >{\raggedright\arraybackslash}p{3.4cm}
    >{\raggedright\arraybackslash}p{4.8cm}
    >{\raggedright\arraybackslash}X
}
\toprule
\textbf{Ref.} &
\textbf{Target crop/material} &
\textbf{Detection task} &
\textbf{Method} &
\textbf{Reported accuracy} \\
\midrule

Nanda \textit{et al.} (2019) &
Wood &
Subterranean termite detection &
LSTM-based deep learning &
97.16\% \\

Escola \textit{et al.} (2020) &
Coffee &
Cicadid pest detection &
SVM &
96.91\% \\

Ali \textit{et al.} (2023) &
Agricultural crops &
Classification of 10 pest species &
CNN--BiLSTM &
98.91\% \\

Ekramirad \textit{et al.} (2023) &
Apples &
Codling moth detection &
AdaBoost &
94.00\% \\

Torky \textit{et al.} (2023) &
Agricultural crops &
Red palm weevil detection &
VGGish, SVM, DT, NB, LR, and DL models &
94.50\% \\

Shetty and Kumar (2025) &
Agricultural crops &
Cicada, beetle, termite, and cricket classification &
XGBoost, Random Forest, KNN &
97.00\% \\

Sedehi \textit{et al.} (2025) &
Agricultural crops &
Subterranean termite detection &
ResNet18, ResNet50 &
95.00\% \\

Wen (2025) &
Agricultural crops &
Four-species pest classification &
CNN, SVM &
98.80\% \\

\bottomrule
\end{tabularx}
\end{table*}

The integration of Artificial Intelligence (AI) with Internet of Things (IoT) technologies enables distributed sensing and continuous agricultural monitoring (Kariyanna \& Sowjanya 2024). IoT-based pest-detection systems typically combine embedded sensors, wireless communication, and edge- or cloud-based processing for real-time field surveillance (Ali \textit{et al.} 2023). However, general audio or visual monitoring systems may lack the sensitivity needed to detect low-amplitude termite acoustic signatures in plantation environments (Mankin \textit{et al.} 2020, 2021). Field deployment is further constrained by environmental noise, limited computational capacity, energy availability, and communication stability on resource-limited edge devices (Samie \textit{et al.} 2016, Bian \textit{et al.} 2022, Folliot \textit{et al.} 2022).

Despite advances in vision-based pest detection and acoustic monitoring, no existing framework specifically addresses the early detection, severity assessment, and spatial mapping of concealed ULWT infestations in large-scale tea plantations. Vision-based methods rely on visible symptoms, while acoustic studies are largely limited to controlled conditions and binary detection. Existing approaches also rarely integrate field acoustic sensing, edge deployment, severity estimation, and geospatial analysis within a single plantation-scale system, limiting real-time, non-destructive ULWT management.

To address this gap, this study develops an integrated AI--IoT acoustic monitoring framework with three objectives: (1) detect concealed ULWT activity using in-field acoustic sensing and edge-based DL; (2) estimate infestation severity from acoustic characteristics; and (3) map infestation distribution using geospatial analytics to support plantation-level monitoring and targeted intervention.

\section{Methodology}

\begin{figure*}[t]
    \centering
    \includegraphics[width=\textwidth]{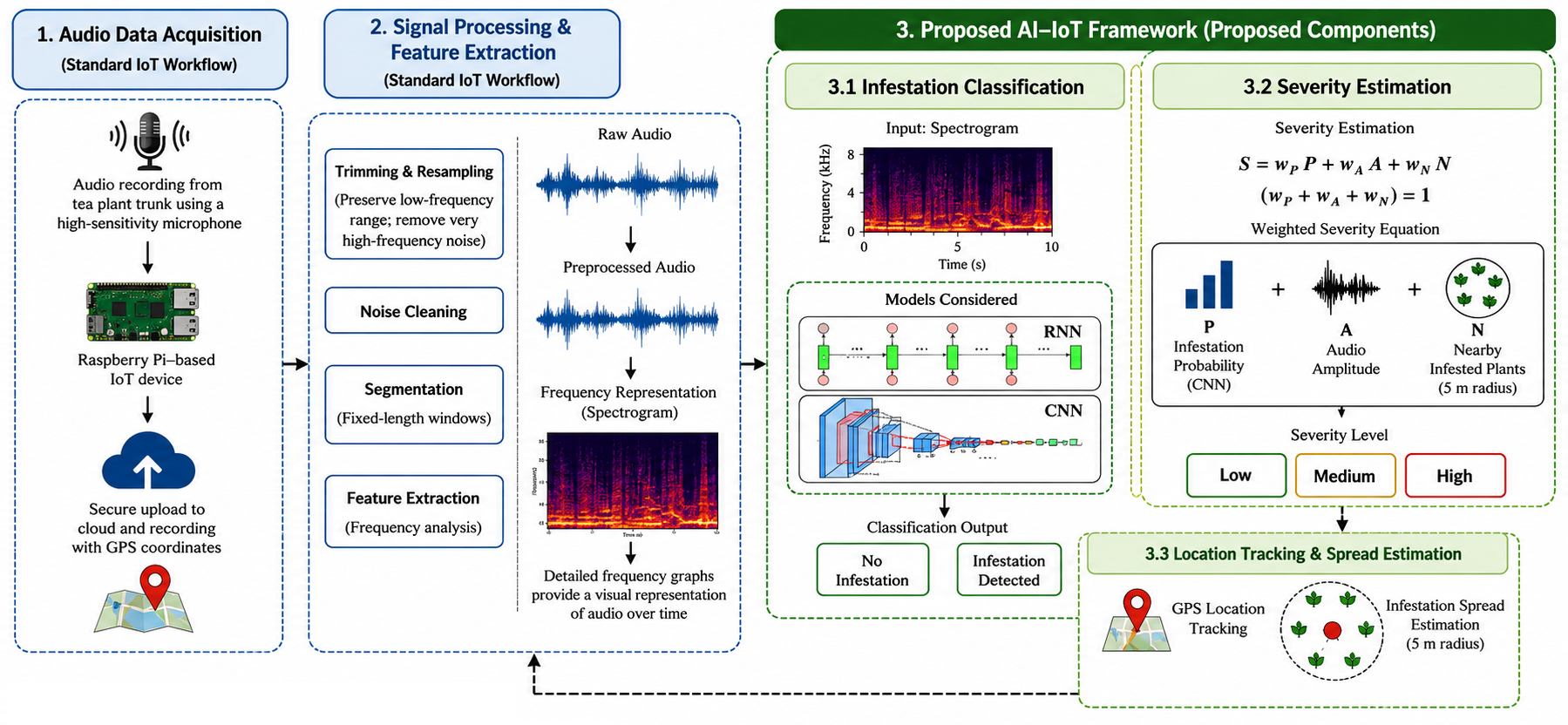}
    \caption{Overview of the proposed AI--IoT framework for acoustic-based ULWT infestation detection in tea plantations, integrating trunk-level acoustic data acquisition and preprocessing, deep learning-based infestation classification, severity estimation, user notification, and geospatial visualization for spread monitoring.}
    \label{fig:proposed_framework}
\end{figure*}

The proposed end-to-end system integrates acoustic data collection from tea trunks, signal preprocessing, infestation classification, severity estimation, user notification, and location-based visualization of potential spread. The framework supports early ULWT detection and monitoring in tea plantations, with the overall workflow shown in Fig.~\ref{fig:proposed_framework}.

Audio data are collected from healthy and infested tea trunks using a high-sensitivity microphone connected to a Raspberry Pi-based IoT device under varying field conditions. Recordings are standardized through trimming and resampling, followed by noise cleaning, segmentation, and feature extraction while preserving the low-frequency components associated with termite activity. Frequency-based representations are then generated for analysis.

The processed signals and frequency representations are analysed using frequency analysis and deep learning. Although RNNs and CNNs were considered, the CNN was selected because of its suitability for extracting local and hierarchical features from spectrogram images. Following classification, the user is notified of the infestation status. For detected infestations, severity is estimated by combining CNN infestation probability, audio amplitude, and the number of nearby infested plants within a 5~m radius using a weighted severity equation. Geographic coordinates are also recorded for spatial tracking.

Finally, the severity value is used to estimate infestation distribution and visualize potential spread within the plantation. By integrating acoustic sensing, classification, severity assessment, and geospatial mapping, the methodology supports timely decision-making and improved ULWT management.

\subsection{Infestation Classification}

\subsubsection{Data Collection}

A dataset of acoustic recordings was constructed to enable the detection of ULWT infestation in tea plants (Senevirathna \textit{et al.} 2026). Audio recordings of 50 seconds in duration were collected from both ULWT-infested tea plant trunks and healthy trunks representing the control class. Data collection was conducted in \textit{Pundaluoya}, a highland area in the \textit{Nuwara Eliya} district of Sri Lanka. Acoustic sampling was conducted within an approximately 3.5~ha plantation area using 10 sampling plots. A total of 40 tea bushes were sampled, including 20 healthy bushes and 20 ULWT-infested bushes. Ten original 50-second acoustic recordings were collected from each tea bush, resulting in 200 original recordings for each class. The tea bushes were randomly selected from the plants available for assessment within the sampling plots with the assistance of personnel from the Tea Research Institute of Sri Lanka (TRISL). The selection was conducted to obtain an equal number of healthy and ULWT-infested bushes while representing different locations within the sampled plantation area.

Recordings were acquired under varying field conditions, including noisy outdoor environments. These conditions helped capture realistic acoustic variability and produced a representative dataset suitable for comparing sound patterns associated with infestation. The infestation status of each sampled tea bush was determined with the assistance of experienced TRISL researchers. Plants assigned to the infested class were identified through expert field inspection and the presence of diagnostic evidence associated with ULWT infestation. Healthy control plants were selected from the same plantation area based on the absence of ULWT infestation evidence during the expert assessment. The same assessment procedure was applied across the 10 sampling plots to maintain consistency in class assignment.

Acoustic recordings were collected under natural plantation conditions rather than under controlled laboratory conditions. Therefore, the recordings incorporated realistic field variability and background interference. The same recording duration, IoT-based acquisition device, and microphone-contact procedure were used for both healthy and infested tea bushes.

To standardize the recordings prior to model development, a preprocessing pipeline was applied. The first and last 5 seconds of each recording were removed to minimize noise introduced during the initiation and termination of the recording process. The trimmed audio files were then resampled to a uniform sampling rate to ensure consistency across all samples.

The increase from 200 to 800 samples per class was achieved through temporal segmentation. Each original 50-second recording was reduced to 40 seconds after trimming and subsequently divided into four non-overlapping 10-second clips. Thus, each original recording generated four segmented audio samples while preserving the low-frequency characteristics associated with termite activity. The original recordings were assigned to the training, validation, and test subsets before temporal segmentation. Segmentation was then performed separately within each subset. Consequently, the 160, 20, and 20 original recordings assigned to the training, validation, and test sets generated 640, 80, and 80 segmented clips, respectively. This procedure prevented clips derived from the same original recording from being distributed across different dataset subsets.

The final dataset composition is presented in Table~\ref{tab:audio_dataset}. As shown in the table, the healthy and infested classes were balanced in terms of the numbers of original and segmented samples. The resulting dataset was used for model training, validation, and performance evaluation.

\begin{table*}[t]
\centering
\caption{Audio dataset composition for ULWT detection.}
\label{tab:audio_dataset}
\renewcommand{\arraystretch}{1.2}

\begin{tabular}{p{2.5cm} p{2.0cm} p{1.6cm} p{1.8cm} c c c c}
\toprule
\textbf{Condition} &
\textbf{Location} &
\textbf{Class} &
\textbf{Type} &
\textbf{Total} &
\textbf{Train} &
\textbf{Val} &
\textbf{Test} \\
\midrule

\multirow{4}{*}{Infested \& Healthy} &
\multirow{4}{*}{\textit{Pundaluoya}} &
\multirow{2}{*}{Healthy} &
Collected &
200 &
160 &
20 &
20 \\

& & &
Segmented &
800 &
640 &
80 &
80 \\
\cmidrule(lr){3-8}

& &
\multirow{2}{*}{Infested} &
Collected &
200 &
160 &
20 &
20 \\

& & &
Segmented &
800 &
640 &
80 &
80 \\

\bottomrule
\end{tabular}
\end{table*}

\subsubsection{Classification Model Selection and Training}

To identify a suitable model for termite infestation detection in tea plantations, multiple deep learning architectures were considered, including feedforward neural networks (FNNs), recurrent neural networks (RNNs), and CNNs. The CNN was selected as the primary classification model because the acoustic recordings were represented as spectrogram images containing localized time--frequency patterns associated with termite activity. CNNs can preserve and learn hierarchical spatial relationships within these image-like inputs (Mushtaq \textit{et al.} 2021, Liyanarachchi \textit{et al.} 2023), whereas FNNs do not explicitly exploit spatial locality and RNNs are primarily designed to model sequential dependencies. Therefore, the CNN was considered the most appropriate architecture for extracting discriminative features from the spectrogram representations used in this study.

In the proposed pipeline, the acoustic recordings were transformed into spectrogram representations using Fourier-based analysis. These spectrograms represent changes in frequency content over time and provide informative patterns associated with termite activity. CNNs are well suited to analyse these representations because they can learn local and hierarchical spatial features from image-based inputs (Mankin 2012). In contrast, FNNs do not explicitly exploit spatial relationships, while RNNs are primarily designed to model sequential dependencies.

The CNN received a single-channel spectrogram image of size $64 \times 64 \times 1$ as input. The architecture consisted of two convolutional layers with 32 and 64 filters, respectively. Each convolutional layer used a $3 \times 3$ kernel, a stride of $1 \times 1$, valid padding, and the ReLU activation function. A $2 \times 2$ max-pooling layer with a stride of $2 \times 2$ followed each convolutional layer to reduce the spatial dimensions of the feature maps.

After the final max-pooling operation, the extracted feature maps were flattened into a vector of 12,544 features and passed through a fully connected layer containing 64 neurons. ReLU activation was applied to the fully connected layer. The output layer contained one neuron with sigmoid activation to generate the probability of the input sample belonging to the infested class. The complete model contained 821,761 trainable parameters.

The model was trained for 50 epochs using a batch size of 32. The Adam optimizer was used with a learning rate of 0.001, $\beta_{1}$ of 0.9, $\beta_{2}$ of 0.999, and $\epsilon$ of $1 \times 10^{-7}$ (Kingma and Ba 2015). Binary cross-entropy was used as the loss function. The model hyperparameters were selected through preliminary empirical trials and commonly adopted settings for binary image classification. The selected configuration was retained based on stable validation convergence and classification performance during model development. Through supervised training, the convolutional layers learned spectral patterns and local variations that distinguish infested tea plants from healthy controls (Mushtaq \textit{et al.} 2021). The resulting model forms the core of the proposed detection framework and supports automated acoustic screening for ULWT infestation in tea plantations.

\subsubsection{IoT-Based Audio Acquisition Device}

An IoT-based audio acquisition device was developed to capture acoustic signals from tea plant trunks for termite infestation detection. The system was built around a Raspberry Pi 3, which served as the central controller for data acquisition, local processing, and communication. A high-sensitivity microphone was connected to the Raspberry Pi to capture low-amplitude acoustic activity from within the tea plant trunks. Similar acoustic sensing approaches have been used for the non-destructive detection of concealed termite activity in trees and other woody structures (Mankin \textit{et al.} 2002), while Raspberry Pi devices have also been employed as low-cost nodes for acoustic acquisition and processing.

During each recording, the microphone was positioned in direct contact with the woody trunk of the selected tea bush using a dedicated fitting interface. This direct-contact arrangement enabled more reliable sound capture from the plant structure while reducing the influence of surrounding environmental noise. The same microphone-placement and trunk-contact procedure was applied to both healthy and infested tea bushes to maintain consistency across the dataset.

The Raspberry Pi coordinated the recording process and managed the transfer of captured audio samples to a cloud-based platform for subsequent storage and analysis (Arce \textit{et al.} 2021, Wanninayake \textit{et al.} 2023). Each recording was captured for 50 seconds using the same device configuration under natural field conditions. Although environmental conditions varied among recording events, the acquisition procedure, recording duration, and microphone-contact method were maintained consistently.

\begin{figure*}[t]
    \centering
    \includegraphics[width=\textwidth]{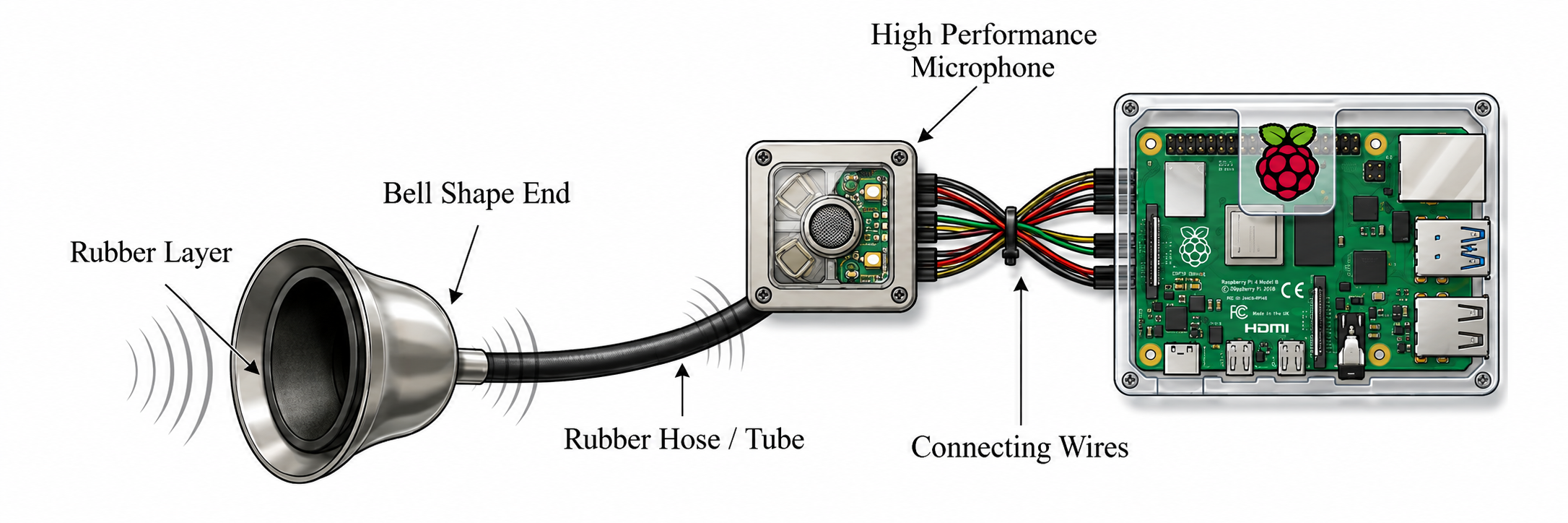}
    \caption{Architecture of the IoT-based audio acquisition device used for capturing acoustic signals from tea trunks.}
    \label{fig:iot_device}
\end{figure*}

In addition to data acquisition, the device supported communication with a mobile application, allowing recordings to be monitored and managed remotely. This integration enabled efficient field deployment and facilitated timely transmission of audio samples for further processing. The overall system, comprising the Raspberry Pi controller, microphone module, trunk interface, cloud connectivity, and mobile application support, provided a practical framework for collecting high-quality acoustic data for deep learning-based termite infestation detection. Fig.~4 illustrates the main components of the proposed IoT device and their integration within the data-acquisition framework. Fig.~\ref{fig:iot_device} illustrates the main components of the proposed IoT device and their integration within the data acquisition framework.

\subsubsection{Severity Estimation of Termite Infestation}

In addition to binary infestation detection, the proposed framework estimates the severity of ULWT infestation in tea plantations using a composite severity score, denoted by $S$. The purpose of this score is to quantify the relative seriousness of infestation by integrating multiple indicators associated with termite activity and local spread. Rather than relying on a single measurement, the severity formulation combines acoustic and spatial factors to provide a more informative assessment of infestation conditions.

The severity score was formulated using three parameters: the infestation probability predicted by the CNN model, the mean amplitude level of the recorded audio signal, and the number of surrounding infested plants within a 5~m radius. These factors were selected because they capture complementary aspects of infestation severity. The infestation probability reflects the confidence of the classification model, the amplitude level provides an additional acoustic descriptor of termite-related activity, and the number of nearby infested plants indicates the degree of local infestation spread. By combining these terms in a weighted formulation, the proposed metric enables the estimation of infestation severity on a continuous scale. The severity score is computed as follows:

\begin{equation}
S = (0.2 \times A \times 10) + (0.3 \times P \times 100) + (0.5 \times N \times 10)
\label{eq:severity}
\end{equation}

where $S$ denotes the severity score, $A$ represents the amplitude level, $P$ is the infestation probability predicted by the CNN model, and $N$ is the number of surrounding infested plants within a 5~m radius. The weighting factors were assigned heuristically based on domain knowledge and expert guidance from TRISL, with greater emphasis placed on the number of surrounding infested plants because spatial clustering was considered a stronger indicator of plantation-level infestation severity. Infestation probability was assigned an intermediate contribution as a model-based indicator, while amplitude was given a lower contribution because it may also be influenced by recording and environmental conditions.

\paragraph{Infestation Probability ($P$).}
Infestation probability ($P$) is obtained from the CNN classifier and represents the predicted likelihood that a tea plant is infested. Derived from learned acoustic patterns, higher $P$ values indicate stronger evidence of infestation and contribute to a higher severity score.

In the severity model, $P$ is assigned a weight of 0.3. This reflects its importance as a model-based indicator while ensuring that severity is not determined solely by classification confidence, as $P$ is combined with signal-level and spatial factors.

\paragraph{Amplitude Level ($A$).}
The mean amplitude level ($A$) represents the intensity of the recorded acoustic signal. Since termite activity can generate characteristic acoustic responses, variations in amplitude may provide useful information about infestation activity (Ghosh \& Samanta 2003). In this study, amplitude complements the CNN-derived infestation probability as an additional acoustic descriptor.

The amplitude term is assigned a weight of 0.2, reflecting its supporting role in severity estimation. While amplitude alone cannot determine infestation severity, it may help distinguish weaker from stronger infestation-related activity.

\paragraph{Number of Surrounding Infested Plants ($N$).}
The number of surrounding infested plants within a 5~m radius ($N$) represents local infestation spread. The 5~m neighbourhood was defined based on expert guidance from the TRISL, and the number of infested plants within this radius was used as an indicator of spatial infestation severity.

Among the three factors, $N$ is assigned the highest weight of 0.5, reflecting the importance of spatial clustering. Higher $N$ values indicate infestation across neighbouring plants, suggesting a more severe and potentially expanding infestation.

\paragraph{Tracking Infested Plant Locations.}
Geographic coordinates of detected infested plants were recorded to support plantation monitoring and management. The associated latitude and longitude data enabled location-aware tracking and spatial visualization of infestation sites.

\subsubsection{Potential Distribution Estimation}

To estimate the potential local spread of ULWT infestation, a distribution metric $D$ was defined based on the severity score $S$. This metric represents the estimated radius of infestation spread around a detected plant and was used to support spatial visualization of infestation extent. Consistent with the severity estimation framework, a maximum reference radius of 5~m was adopted according to expert guidance. The severity score was linearly mapped to this radius such that a severity value of 100 corresponds to a distribution radius of 5~m. The potential distribution radius was therefore computed as

\begin{equation}
D = S \times 0.05
\label{eq:distribution}
\end{equation}

where $D$ denotes the estimated distribution radius in metres and $S$ represents the severity score. This linear formulation provides a simple and interpretable approximation for translating severity into local spread radius. Higher values of $D$ indicate a greater potential extent of infestation around the detected plant.

\section{Results and Discussion}

\subsection{Results}

\paragraph{IoT Device.}

\begin{figure*}[t]
    \centering

    \begin{subfigure}[b]{0.48\textwidth}
        \centering
        \includegraphics[height=6cm,keepaspectratio]{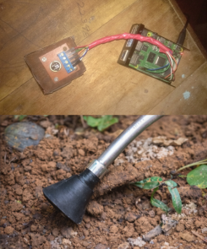}
        \caption{}
        \label{fig:iot_internal}
    \end{subfigure}
    \hspace{0.5cm}
    \begin{subfigure}[b]{0.48\textwidth}
        \centering
        \includegraphics[height=6cm,keepaspectratio]{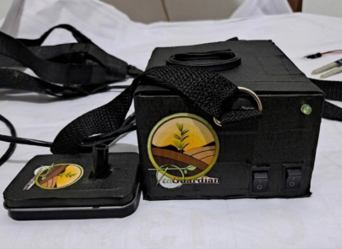}
        \caption{}
        \label{fig:iot_prototype}
    \end{subfigure}

    \caption{Internal components (A) and assembled prototype (B) of the proposed IoT device for acoustic data acquisition.}
    \label{fig:iot_device_prototype}
\end{figure*}

The proposed IoT device collected acoustic recordings from tea trunks in ULWT-affected plantation environments in \textit{Pundaluoya}. Its specialized trunk-fitting interface reduced surrounding environmental noise, improving the quality and consistency of captured signals, while cloud-enabled transmission supported efficient data transfer and storage. The device components and assembled prototype are shown in Fig.~\ref{fig:iot_prototype}.

Field deployment demonstrated the device's suitability for acoustic monitoring under real plantation conditions. The collected recordings were subsequently used as input to the deep learning pipeline for infestation classification, confirming the device's effectiveness as an acquisition platform for capturing ULWT-related signals.

From an application perspective, the developed IoT device establishes a practical basis for integrating acoustic sensing with automated infestation analysis. Its ability to support field data collection, remote transfer, and downstream processing suggests that it may be useful for scalable monitoring of termite infestation in tea plantations.

\begin{figure*}[t]
    \centering

    \begin{subfigure}[b]{0.48\textwidth}
        \centering
        \includegraphics[height=6cm,keepaspectratio]{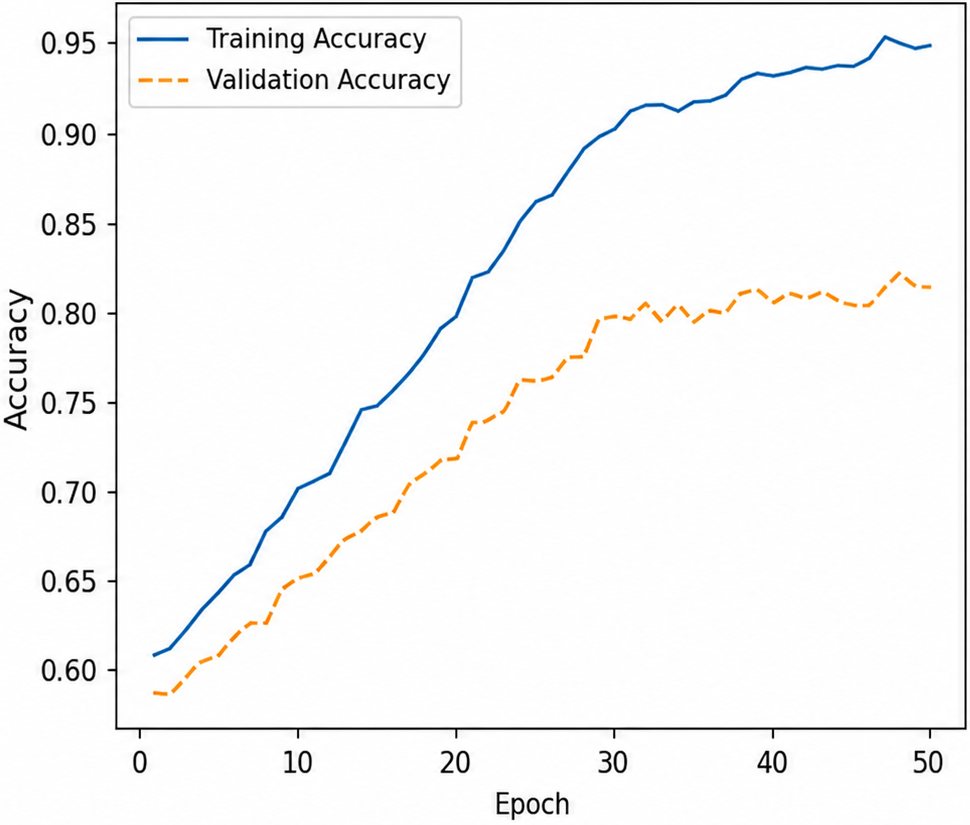}
        \caption{}
        \label{fig:cnn_accuracy}
    \end{subfigure}
    \hspace{0.3cm}
    \begin{subfigure}[b]{0.48\textwidth}
        \centering
        \includegraphics[height=6cm,keepaspectratio]{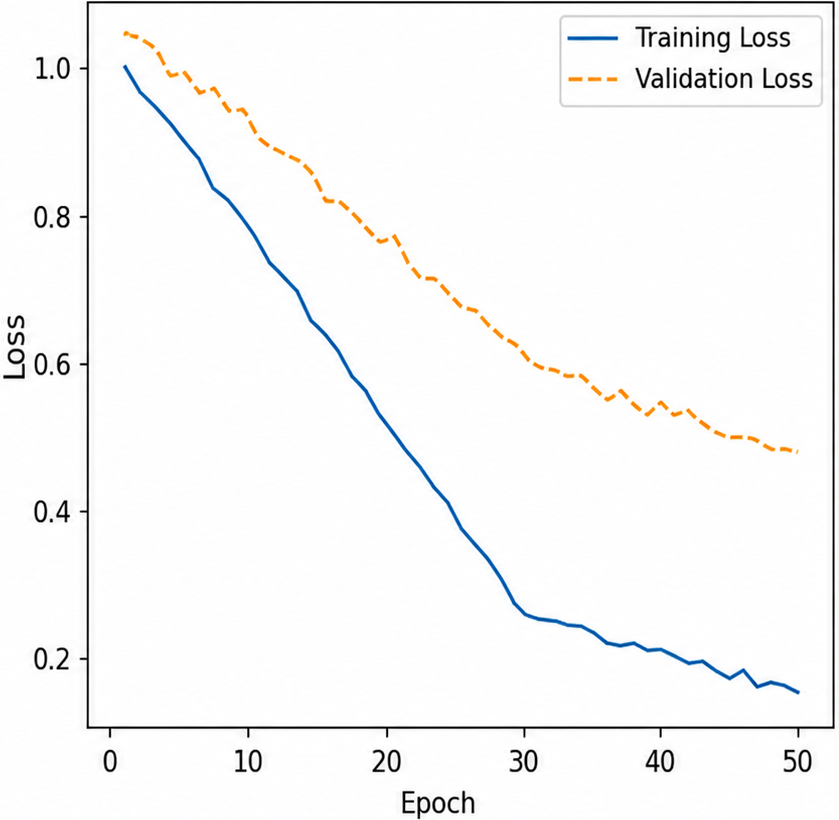}
        \caption{}
        \label{fig:cnn_loss}
    \end{subfigure}

    \caption{Training and validation performance curves of the CNN-based ULWT infestation classification model using Fourier-derived spectrogram representations: (A) accuracy and (B) loss.}
    \label{fig:cnn_training}
\end{figure*}

\paragraph{Deep Learning Classification Model.}
The CNN-based classification model achieved an overall accuracy of 81.5\% on the held-out test set. The training and validation accuracy and loss curves over 50 epochs are presented in  Fig.~\ref{fig:cnn_training}. Training accuracy increased from approximately 61\% in the first epoch to 95\% by epoch 50, while validation accuracy increased progressively and stabilised at approximately 81\% after epoch 30. Similarly, training loss decreased from approximately 1.00 to 0.17, whereas validation loss decreased from approximately 1.05 to 0.48.

The difference between the final training and validation accuracies indicates that the model fitted the training data more closely than the validation data. However, the validation accuracy and loss remained relatively stable during the later training epochs, with no substantial deterioration in validation performance.

\begin{figure*}[t]
    \centering

    \begin{subfigure}[b]{0.48\textwidth}
        \centering
        \includegraphics[height=6cm,keepaspectratio]{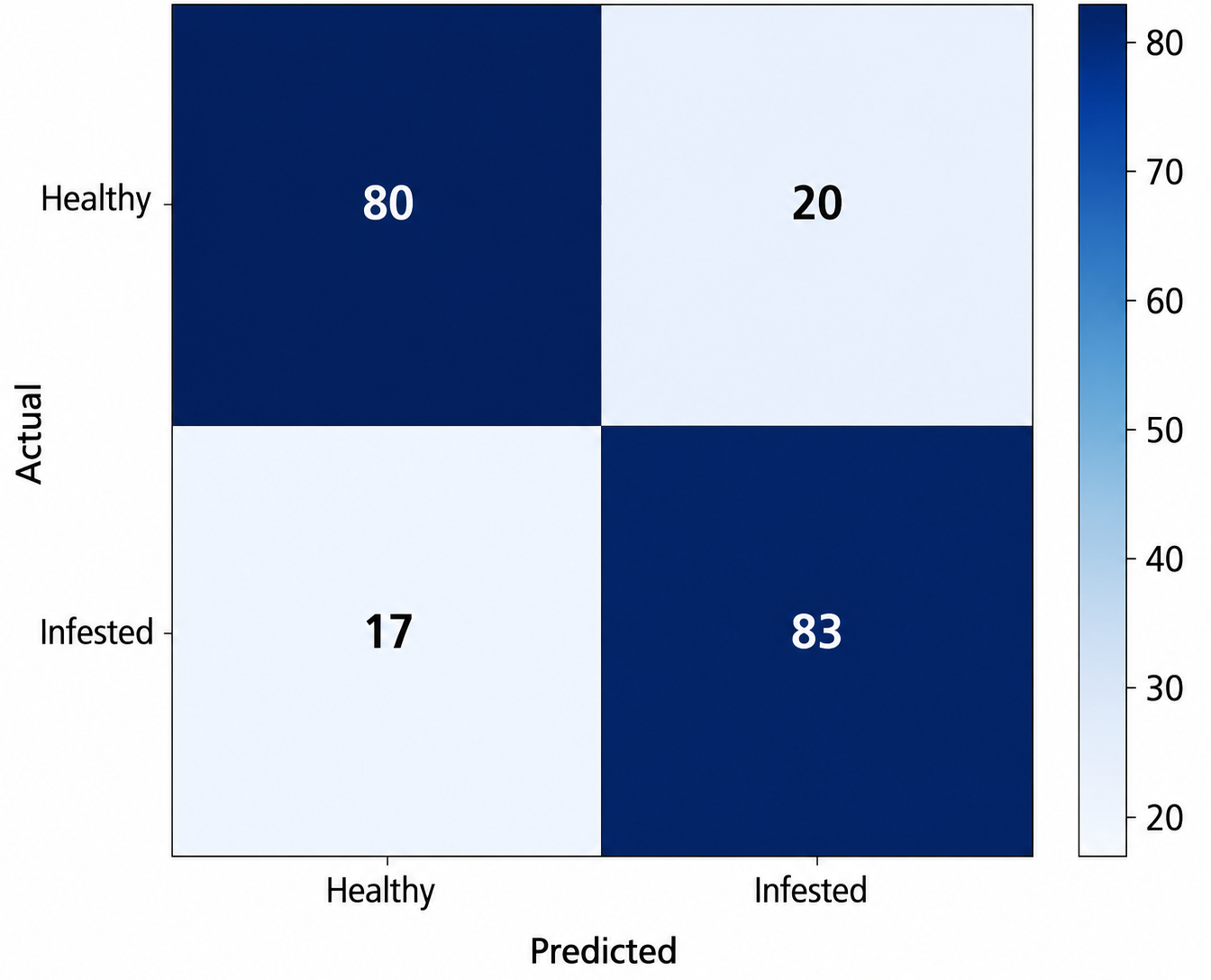}
        \caption{}
        \label{fig:roc_curve}
    \end{subfigure}
    \hspace{0.3cm}
    \begin{subfigure}[b]{0.48\textwidth}
        \centering
        \includegraphics[height=6cm,keepaspectratio]{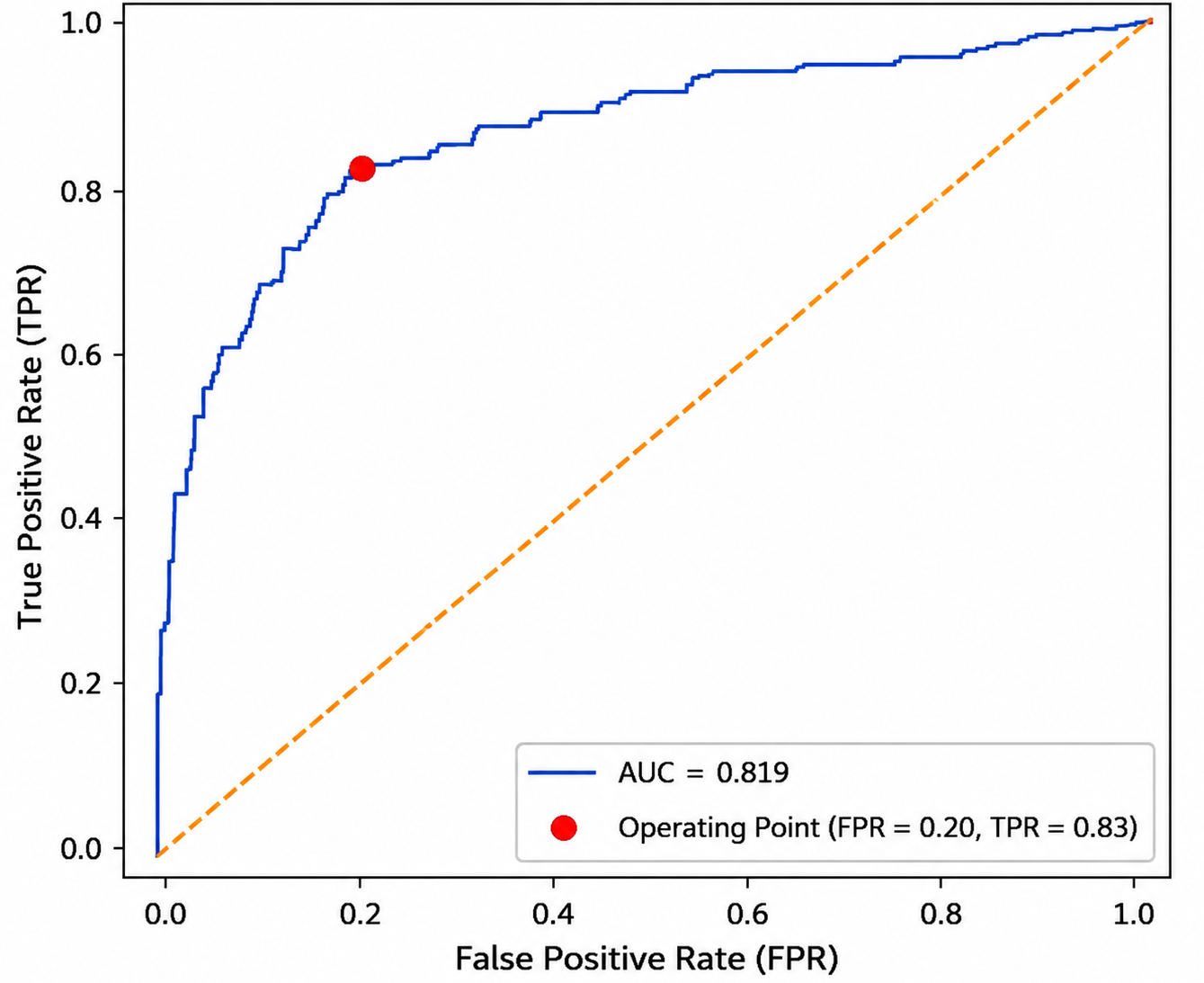}
        \caption{}
        \label{fig:confusion_matrix}
    \end{subfigure}

    \caption{Performance evaluation of the CNN-based ULWT infestation classification model on the held-out test set: (A) ROC curve and (B) confusion matrix.}
    \label{fig:model_evaluation}
\end{figure*}

The discriminative performance of the model was further evaluated using the Receiver Operating Characteristic (ROC) curve shown in Fig.~\ref{fig:roc_curve}. The model achieved an Area Under the Curve (AUC) of 0.819, indicating that it could distinguish between infested and healthy samples across a range of classification thresholds. At the selected operating threshold, the model achieved a false-positive rate of 0.20 and a true-positive rate of 0.83.

The confusion matrix shown in Fig.~\ref{fig:confusion_matrix} provides a sample-level summary of the classification results. The model correctly classified 83 infested samples and 80 healthy samples. In contrast, 20 healthy samples were misclassified as infested, while 17 infested samples were misclassified as healthy. These results corresponded to a precision of 80.6\%, recall of 83.0\%, and F1-score of 0.818 for the infested class.

Overall, the test-set results demonstrate that the CNN could distinguish between ULWT-infested and healthy acoustic samples using Fourier-derived spectrogram representations.

\subsection{Discussion}

The findings demonstrate that acoustic recordings collected from tea plant trunks contain discriminative information that can support the identification of ULWT infestation. The CNN achieved a test accuracy of 81.5\%, an AUC of 0.819, a precision of 80.6\%, a recall of 83.0\%, and an F1-score of 0.818 for the infested class. Rather than considering accuracy alone, the combined performance metrics indicate that Fourier-derived spectrograms captured acoustic patterns associated with infestation, while also showing that reliable discrimination under field conditions remains challenging.

The recall of 83.0\% is particularly important for field screening because it indicates that the model correctly identified 83 of the 100 infested test samples. However, the remaining 17 infested samples were incorrectly classified as healthy. These false-negative predictions are operationally more important than false positives because concealed infestations may remain undetected and continue to damage apparently healthy tea plants. In contrast, the 20 false-positive predictions would mainly result in additional field inspection or confirmatory assessment. Future model optimization should therefore prioritise reducing false negatives while maintaining an acceptable false-positive rate. Potential approaches include classification-threshold optimization, cost-sensitive learning, improved noise filtering, and ensemble classification.

The difference between the final training accuracy of approximately 95\% and the validation and test accuracies of approximately 81\% suggests that the model learned some training-specific patterns that did not transfer fully to unseen samples. Although the validation accuracy and loss remained stable after approximately 30 epochs, the performance gap indicates some degree of overfitting or sensitivity to acoustic characteristics that were not sufficiently represented during training. These characteristics may include variation in tea plant structure, colony size, colony position within the trunk, termite activity level, sensor contact, and environmental noise.

The test accuracy obtained in the present study was lower than the values reported in several previous acoustic pest-detection studies. Earlier studies have reported accuracies of approximately 96.91\% for acoustic detection of cicadid pests in coffee plants, 97.16\% for subterranean-termite detection using combined acoustic and temperature features, 94.0\% for codling moth detection, 94.5\% for red palm weevil detection, 95.0\% for subterranean-termite classification using a ResNet-based model, and 98.91\% for multi-species pest classification (Ali \textit{et al.} 2023). Compared with the reported range of 94.0--98.91\%, the accuracy of 81.5\% achieved in the present study was approximately 12.5--17.4 percentage points lower.

However, these quantitative differences should not be interpreted as a direct ranking of the detection methods because the studies differed substantially in target species, sensing equipment, datasets, input representations, environmental conditions, and evaluation procedures. Several previous studies used controlled insect populations, prepared wood specimens, laboratory chambers, or comparatively constrained acoustic environments. Some also combined acoustic measurements with temperature information or used handcrafted acoustic features. In contrast, the present study used recordings acquired non-destructively from naturally infested living tea plants under operational plantation conditions.

The field-acquired recordings may therefore have been affected by wind, soil organisms, plantation activities, variation in trunk structure, sensor attachment, and differences in termite colony activity. The lower accuracy may consequently reflect the greater variability and complexity of the field environment rather than an inherent limitation of the CNN architecture alone. This distinction is important because models developed under controlled conditions may not necessarily retain the same performance when deployed in operational plantations.

Differences in evaluation design also limit direct comparisons between studies. Reported accuracy can be affected by the number of independent biological samples, the method used to divide the data, the duration and segmentation of recordings, and whether clips from the same plant or original recording are distributed across different dataset partitions. Therefore, numerical performance values from different studies should be compared cautiously unless similar acquisition and validation protocols are used.

A limitation of the present study is the relatively small number of independent field recordings. Although audio segmentation increased the number of model inputs, it did not increase the number of independent tea plants, termite colonies, plantation locations, or recording events. Segments originating from the same recording may share similar infestation-related and background characteristics. Consequently, the number of segmented audio clips should not be interpreted as equivalent to the number of independent biological samples. The model was also evaluated using a single training, validation, and test partition. The reported performance may therefore have been influenced by the particular sample allocation. Repeated hold-out evaluation or plant-level \textit{k}-fold cross-validation would provide a more robust estimate of performance variability and generalization. All segments originating from the same recording or tea plant should remain within the same data partition to prevent information leakage between training and testing.

Environmental parameters were not quantitatively documented during acoustic data collection. Although the trunk-fitting interface was designed to reduce surrounding noise, factors such as wind speed, rainfall, temperature, background-noise intensity, soil-organism activity, and sensor contact pressure may still have affected the recordings. Without these measurements, the influence of individual environmental variables on correct and incorrect classifications could not be assessed. Future work should record such conditions alongside the acoustic data to determine how model performance changes across different field environments.

Despite these limitations, the contribution of the study extends beyond the CNN accuracy alone. The proposed framework integrates trunk-mounted acoustic sensing, cloud-enabled data transfer, deep learning-based infestation classification, severity estimation, and geographic data collection. Previous acoustic studies have commonly concentrated on detecting insect activity or classifying infestation status, whereas the present framework connects acoustic detection with severity assessment and plantation-level spatial monitoring.

From a plantation-management perspective, the proposed framework could support a targeted inspection strategy by identifying tea plants or plantation areas with a higher likelihood of concealed ULWT activity. Acoustic predictions, severity estimates, and geographic information could be used together to prioritise plants for confirmatory inspection and treatment rather than relying only on visible symptoms or broad plantation-wide intervention. This may help direct limited labour and management resources towards locations requiring more immediate attention. However, the system should currently be regarded as a screening and decision-support tool rather than a replacement for expert field assessment.

The findings should therefore be interpreted as an initial field-based demonstration of the feasibility of non-destructive ULWT detection rather than a definitive estimate of performance across all tea-growing environments. Further validation should include larger numbers of independent plants and termite colonies, multiple plantation locations, seasonal data, documented environmental conditions, and repeated plant-level evaluation. These developments would establish whether the proposed framework can achieve the generalisability, sensitivity, and operational reliability required for routine plantation surveillance.

\section{Conclusions}

This study presented an integrated AI--IoT framework for the non-destructive detection and monitoring of ULWT infestation in tea plantations. The framework combined a trunk-mounted acoustic acquisition device, a mobile application for field data collection, a CNN-based classifier, severity estimation, and web-based geospatial visualization within a unified monitoring system. On the held-out test set, the CNN achieved an accuracy of 81.5\%, an AUC of 0.819, a precision of 80.6\%, a recall of 83.0\%, and an F1-score of 81.8\%. These results demonstrate the feasibility of acoustic-based ULWT detection under field conditions.

The main contribution of the study extends beyond binary infestation classification. By integrating infestation probability, acoustic amplitude, and the number of nearby infested plants, the proposed framework provides an initial approach for estimating infestation severity and visualizing its potential spatial distribution. This integration will support plantation managers in identifying higher-risk areas, prioritising confirmatory inspections, and implementing more targeted field interventions. However, due to the occurrence of false-negative predictions, the system should currently be considered a screening and decision-support tool rather than a replacement for expert field inspection. The severity weights and the relationship used to estimate potential distribution were heuristically defined and were not independently validated against measured infestation intensity or observed spread. The main limitations of the study include the relatively small number of independent tea plants and recording events, the absence of recorded environmental variables, and the use of a single dataset partition for evaluation.

Future work will focus on expanding the dataset under a broader range of plantation and environmental conditions, evaluating alternative deep learning architectures, and further validating the proposed severity estimation model. These developments are expected to improve the robustness and generalisability of the framework for wider plantation-level application.

\section*{Acknowledgements}

The authors gratefully acknowledge the Tea Research Institute of Sri Lanka (TRISL) for its expert knowledge, technical guidance, and domain expertise throughout this study.

\balance

\end{document}